\documentclass[10pt,twocolumn,letterpaper]{article}

\usepackage{report}              % To produce the CAMERA-READY version

\usepackage{multirow}

\usepackage{booktabs}
\usepackage{bbding}
\def\etal{{\em et al.}}

\usepackage[colorlinks=true, linkcolor=blue, citecolor=blue, urlcolor=blue]{hyperref}

\begin{document}

\title{Synthetic Data in AI: Challenges, Applications, and Ethical Implications}  % **** Enter the paper title here

\author{Shuang Hao, Wenfeng Han, Tao Jiang, Yiping Li, Haonan Wu,\\ Chunlin Zhong, Zhangjun Zhou, He Tang\thanks{Corresponding author: \texttt{hetang@hust.edu.cn}}  \\ \\
School of Software Engineering, Huazhong University of Science and Technology\\
{\tt\small hetang@hust.edu.cn}
}
% \and
% \and
\maketitle
\section{Introduction}
In the rapidly evolving field of artificial intelligence, the creation and utilization of synthetic datasets have become increasingly significant. This report delves into the multifaceted aspects of synthetic data, particularly emphasizing the challenges and potential biases these datasets may harbor. It explores the methodologies behind synthetic data generation, spanning traditional statistical models to advanced deep learning techniques, and examines their applications across diverse domains. The report also critically addresses the ethical considerations and legal implications associated with synthetic datasets, highlighting the urgent need for mechanisms to ensure fairness, mitigate biases, and uphold ethical standards in AI development.
\section{The generation of synthetic data}
% Real data comes from the real world, and the source of synthetic data is real data generated by mathematical models or algorithms. Because real data collected in the real world is often incomplete and limited, generating data to eliminate discrimination has been widely needed. This demand has also led to the emergence of data synthesis methods. Synthetic data generator can take many forms, from traditional statistical models to modern deep learning models, such as Variational Auto-Encoder(VAE), Generative Adversarial Network(GAN), Diffusion Models and Large Language Models(LLMs). It is important to note that while these methods have had some success, synthetic data is still not a true substitute for real data.
Real data typically refers to data collected directly from the real world, covering text, images, video, audio and so on. However, due to its inherent limitations and incompleteness, issues such as data imbalance\cite{thabtah2020data} and data discrimination\cite{favaretto2019big} arise in practical applications. Since it is difficult to satisfy the demand relying solely on real data, researchers start to employ diverse methods for generating synthetic data through existing real data. These methods range from traditional statistical models to contemporary advanced techniques based on deep learning models.
\subsection{Statistical Models}
% Synthesizing data based on statistical methods is the process of generating new synthetic data by using statistical properties of existing real data. These methods rely on the analysis of the distribution, relationships, and characteristics of real data and attempt to generate synthetic data with similar properties by simulating these statistical properties. The main methods are summarized as follows:
Statistical Models for generating synthetic datasets often rely on the analysis of the distribution, relationships, and characteristics of real data and attempt to generate synthetic data with similar properties by simulating these statistical properties. The main methods are summarized as follows:
\\\noindent{}\textbf{Distribution-based methods.} This method aims to simulate the distribution characteristics of the original data. For continuous data, probability density functions (PDFs) are used to describe the distribution, while probability mass functions (PMFs) are used for discrete data. When synthesizing data, new data points can be generated by sampling from the distribution of existing data.
\\\noindent{}\textbf{Interpolation and Extrapolation.} Interpolation and extrapolation involve generating new data points between or beyond existing data points. This is particularly useful for time series, geographical data, etc. One common interpolation method is linear interpolation, where the value of a new point is determined by a linear relationship between two adjacent known points.
\\\noindent{}\textbf{Monte Carlo Simulation.} Monte Carlo simulation employs random sampling to simulate the uncertainty in real systems. In data synthesis, this method is utilized to generate new samples by randomly sampling from known distributions. It finds common applications in finance, engineering, and physics modeling.
\\\noindent{}\textbf{Model-based Sampling.} This method involves utilizing statistical models of existing data to predict new data points. For example, a linear regression model can be fitted to existing data, and new data points can be generated by randomly sampling the model parameters. This approach is particularly effective for data exhibiting linear relationships.
\\\noindent{}\textbf{Kernel Density Estimation.} Kernel density estimation involves placing kernels (typically Gaussian kernels) around each data point and calculating the contribution at each point to estimate the probability density function. When generating new samples, random sampling can be performed from the estimated probability density function. This is useful for capturing the complexity and multimodality of data distributions.

\subsection{Deep learning based generation}
In light of the rapid advancements in deep learning over recent years, scholars have increasingly directed their attention toward harnessing deep learning methodologies for the generation of synthetic datasets. In contrast to traditional statistical-based methods, deep learning approaches have the capability to acquire more intricate feature representations of data without the need for manually designing feature extractors. Their inherent nonlinearity makes them well-suited for adapting to the complex nonlinear relationships within data, enabling a more effective capture of the essential characteristics in the data. 
The following section outlines several typical approaches to data generation based on deep learning.
\\\noindent{}\textbf{Variational Auto-Encoder(VAE).}
VAE\cite{kingma2013auto} is a kind of probabilistic generative model, which employs the encoder-decoder architecture. The encoder maps the input data to the underlying latent space corresponding to the parameters of the variational distribution and the decoder projects features from the latent space back into the input space. By capturing the distribution of latent space features, the VAE can generate multiple distinct samples that follow the same distribution.The inherent randomness of the VAE introduces a degree of diversity in the generated data, making them more representative of the complexity in real datasets.
\\\noindent{}\textbf{Generative Adversarial Networks(GAN).}
GAN was first proposed by Goodfellow \etal\cite{goodfellow2020generative} in 2014, which consists of a generator and a discriminator. The generator randomly samples from the latent space, in order to produce samples resembling the training set. Meanwhile, the discriminator whether the sample belongs to real or synthetic data. These two parts engages in a continual adversarial learning process and updates parameters alternately until the generator is able to synthesise high quality samples to fool the discriminator. 
\\\noindent{}\textbf{Diffusion models.}
Diffusion model\cite{ho2020denoising} stands as a robust method for crafting synthetic datasets, relying on a systematic approach to emulate intricate temporal dependencies within data. Rooted in the principles of diffusion processes, where information spreads through a network over time, this technique integrates these models into the data generation process. By doing so, it enables the replication of nuanced patterns and relationships observed in authentic data. The crux lies in accurately capturing the dynamic evolution of information over time. This technique, grounded in diffusion models, not only reproduces statistical characteristics but also adeptly mirrors the complex temporal dynamics present in real-world datasets. This algorithmic approach proves invaluable for generating synthetic datasets, offering a sophisticated tool for simulating realistic data scenarios essential for the training and evaluation of machine learning models across diverse domains. To date, synthetic data generated from diffusion models has found extensive application across various vision tasks, notably enhancing performance in areas such as image classification\cite{azizi2023synthetic}, object detection\cite{Fang2024}, and semantic segmentation\cite{nguyen2023dataset}.
\\\noindent{}\textbf{Large language models.}                                                                 
In the past years, large language models (LLMs) have emerged as a revolutionary approach for generating synthetic datasets.
For instance, models such as GPT-3.5, with their exceptional in-context learning capabilities and extensive pre-trained linguistic knowledge, exemplify the capacity of large language models (LLMs) to produce synthetic datasets. This capability facilitates the training of models on smaller domains, effectively addressing the challenge of data scarcity in specific areas.
The Generative Pre-trained Transformer (GPT) family\cite{radford2018improving,radford2019language,brown2020language} comprises a series of Natural Language Processing (NLP) models developed by OpenAI, employing the Transformer architecture to capture long-distance dependencies in input sequences. GPT is unsupervisedly pre-trained on large-scale Internet text corpora, for example, to predict the next word in a given context. This pre-training strategy enables the model to acquire a profound understanding of linguistic statistical structure and contextual relationships and ultimately to perform a variety of natural language processing tasks in a pre-training-fine-tuning format.
For example, Josifoski \etal\cite{josifoski-etal-2023-exploiting} synthetically generate a dataset of 1.8M data points in a reverse manner and demonstrate the effectiveness of this approach on closed information extraction. Whitehouse \etal\cite{whitehouse-etal-2023-llm} utilise several LLMs, namely Dolly-v2, StableVicuna, ChatGPT, and GPT-4 to augment three datasets and assess the naturalness and logical coherence of the generated examples.

\section{The usage of synthetic data.}
\subsection{Existing synthetic datasets.}
Synthetic data possesses several advantages that natural data lacks, making it an attractive choice in many fields. Compared to natural data, synthetic datasets are relatively easy to acquire and can provide data in rare or challenging scenarios, thereby addressing diversity issues in certain datasets. Additionally, this technology can effectively avoid privacy concerns, safeguarding user information security. As this technology gradually becomes more prominent, its practical applications are becoming increasingly widespread. This section will discuss several domains where generated data has had a significant impact.
\\\noindent{}\textbf{Vision}
Synthetic datasets can greatly address the challenge of acquiring natural data in certain domains. In the early days of computer vision, the generation of corresponding datasets relied primarily on computer graphics engines\cite{sun2019dissecting,yao2020simulating,tang2019pamtri,wang2019learning}. For example, in the re-identification (Re-ID) domain, Sun \etal\cite{sun2019dissecting} created the PersonX dataset using the a engine based on Unity. This dataset includes three pure color backgrounds and three scene backgrounds. It consists of 1266 hand-crafted identities (547 females and 719 males), with each identity having 36 viewpoints. At that time, this dataset effectively addressed the lack of multi-viewpoint data. Furthermore\cite{zhu2017unpaired,torkzadehmahani2019dp,brock2018large,niemeyer2021giraffe}, there are also generated methods based on Generative Adversarial Networks (GAN). In \cite{niemeyer2021giraffe}, GIRAFFE introduces synthetic neural feature maps, enabling control over camera poses, object placements and orientations, as well as object shapes and appearances during generation. Moreover, GIRAFFE allows for the free addition of multiple objects in a scene, expanding the generated scenes from single-object to multi-object scenarios.
\\\noindent{}\textbf{Audio}
Synthetic data is widely employed in the field of audio, and its rapid development is truly remarkable. Take, for instance, the Speech Commands Dataset proposed by Chris Donahue \etal, which leverages WaveGAN, a generative adversarial network\cite{donahue2018adversarial}. WaveGAN excels in synthesizing one-second audio waveform slices with global coherence, making it particularly well-suited for sound effects generation. Even without labels, when trained on small vocabulary speech datasets, WaveGAN adeptly learns to generate intelligible words and can extend its synthesis capabilities to audio from diverse domains, including drums, bird sounds, and piano.Furthermore, Zhang \etal introduced Stutter-TTS\cite{zhang2022stutter}, tailored to tackle the performance challenges faced by existing Automatic Speech Recognition (ASR) interfaces when dealing with stuttered speech. Stutter-TTS stands as an end-to-end neural text-to-speech model with the proficiency to synthesize various forms of stuttered speech. It employs a simple yet effective prosody control strategy and incorporates additional markers during training to represent specific stuttering features. By strategically selecting the positions of these markers, Stutter-TTS provides word-level control over the occurrence of stuttering in the synthesized speech.
\\\noindent{}\textbf{Natural Language Processing (NLP)}
The growing interest in synthetic data has propelled the thriving development of numerous deep generative models in the field of Natural Language Processing (NLP)\cite{dewi2022synthetic}. In recent years, machine learning has demonstrated its formidable capabilities in tasks such as classification, routing, filtering, and information retrieval across various domains\cite{forman2003extensive}.Addressing the challenge of synonym variations arising from contextual changes in NLP, researchers have introduced the BLEURT model. Built upon BERT, this model simulates human judgment by utilizing a limited set of training examples that may exhibit biases. To enhance the model's generalization, innovative pre-training approaches have been developed using millions of synthetic examples\cite{yue2022synthetic, zheng2021using}.Furthermore, RelGAN, developed by Rice University, represents a significant breakthrough in text generation using Generative Adversarial Networks (GAN). Comprising three key components - a relation memory-based generator, Gumbel-Softmax relaxation algorithm, and multiple embedded representations in the discriminator - RelGAN outperforms several state-of-the-art models in benchmark tests, showcasing its remarkable performance in sample quality and diversity. This underscores its potential for further research and application across a wide range of NLP tasks and challenges\cite{nie2018relgan,zhao2022act}.
\\\noindent{}\textbf{Health}
In the realm of healthcare, the generation of synthetic data plays a pivotal role in comprehending diseases while upholding patient confidentiality and privacy \cite{2019SynSys}. Synthetic data possesses the capability to mirror the original data distribution without disclosing actual patient information \cite{2019SynSys,2019Integrated,2021Generating}. A notable example in healthcare is MedGAN, a model introduced by Edward Choi \etal, leveraging adversarial networks to generate authentic synchronized medical records. Through the integration of autoencoders and generative adversarial networks, MedGAN proficiently produces high-dimensional discrete variables (e.g., binary features and counting features) based on genuine medical records [16]. Synthetic patient records generated by MedGAN have demonstrated comparable performance to real data across various experiments, encompassing distribution statistics, predictive modeling tasks, and assessments by medical experts. Furthermore, synthetic data finds widespread application in the realm of drug discovery. The predominant approach involves learning the distribution of drug molecules from existing databases and subsequently deriving new samples (i.e., drug molecules) from the acquired knowledge of drug molecule distributions. Numerous implementations of this process exist, such as variant autoencoders (VAEs)\cite{2018Junction,gomez2018automatic,zhang2021ddn2}, generative adversarial networks (GANs) \cite{de2019molgan}, energy-based models (EBMs) \cite{fu2022antibody,fu2021differentiable}, diffusion models \cite{xu2022geodiff}, reinforcement learning (RL) \cite{zhou2019optimization,olivecrona2017molecular,fu2022reinforced}, genetic algorithms \cite{jensen2019graph}, sampling-based methods \cite{fu2021mimosa,fu2022sipf}, and others.

\begin{table*}[t]
\centering
\fontsize{9}{10}\selectfont
\caption{Summarization of Representative Works in Synthetic Data Generation.}
\begin{tabular}{llllll}
\toprule
Datasets& Generated Method& Applicantion &Size  &Content&DL  \\ 
\midrule
\multirow{1}{*}{Kubric}\cite{kundu2022kubric} & 3D-Rendered & Vison &—— & 3D Image/Video &\Checkmark  \\
\multirow{1}{*}{Structured3D}\cite{zheng2020structured3d} & 3D-Rendered & Vison &196,515 Frames & 3D Image/Video& \Checkmark  \\
\midrule
\multirow{1}{*}{PersonX}\cite{sun2019dissecting} & Physically Realistic Engines & Vison &1266 Images & Person image &\XSolid\\ 
\multirow{1}{*}{GCC}\cite{wang2019learning} & Physically Realistic Engines & Vison &15,212 Images & Crowd image &\XSolid
\\
\midrule
\multirow{1}{*}{BigGANs}\cite{gowal2021improving} & GAN & Vison &—— & Image Annotation &\Checkmark
\\
\multirow{1}{*}{GIRAFFE}\cite{niemeyer2021giraffe} & GAN & Vison &—— & Mutil-View image &\Checkmark
\\\midrule
\multirow{1}{*}{SyntheticData}\cite{he2022synthetic} &Diffusion Model & Vison &—— & Generate Rare Species Data &\Checkmark\\
\multirow{1}{*}{}\cite{he2022synthetic} &Diffusion Model & Vison &—— & Improved Imagenet &\Checkmark
\\\midrule
\multirow{1}{*}{SynthIE}\cite{josifoski-etal-2023-exploiting} &LLM & NLP &1.8M data
points& Text &\Checkmark\\
\multirow{1}{*}{Gen-X}\cite{whitehouse-etal-2023-llm} &LLM & NLP &——& Data Augmentation&\Checkmark\\
\bottomrule
\end{tabular}
\label{table_LQA Comparison}
\end{table*}

% With the development of Artificial Intelligence Generated Content (AIGC), most existing synthetic datasets are now generated based on visual-language models. For example, methods like stable diffusion or midjourney can generate corresponding images based on text descriptions. This further aligns synthetic datasets with natural datasets.

% However, synthetic datasets also face many issues when used. Ensuring the effectiveness and diversity of generated data is an inherent challenge of synthetic datasets. Since the training data for image generation models often cannot guarantee the fairness of data sources, synthetic datasets tend to exhibit existing biases. For example, when using positive terms like "rich," "productive," or "smart" without specifying a particular race, the generated models often depict white individuals. On the other hand, negative terms like "poor" or "social service" frequently result in the depiction of black or Asian individuals. This also leads to similar biases in subsequent synthetic datasets based on these diffusion models.
\subsection{The data distribution of synthetic datasets.}
In the rapid advancement of artificial intelligence and machine learning, synthetic datasets have become an essential resource. These datasets, typically algorithmically generated, are used for training and testing a variety of models. However, the generation process of synthetic datasets often harbors implicit issues, especially regarding the fairness and representativeness of data distribution. These issues can affect the performance of models and potentially lead to biases and discriminatory practices in real-world applications.
\\\noindent{}\textbf{Distribution Issues in Synthetic Datasets.}
The generation of synthetic datasets often lacks sufficient consideration for demographic diversity, which can lead to unbalanced data distributions in terms of gender, age, race, etc. For instance, in the creation of datasets, if the data is primarily based on individuals from specific racial or age groups, the trained models might perform poorly when dealing with other groups. This situation can lead to severe discriminatory issues in real-world applications, such as certain racial groups being incorrectly identified or entirely overlooked. For instance, if a dataset used to train a facial recognition system disproportionately represents certain demographics over others, the resulting AI models may exhibit biased performance, leading to unfair or discriminatory outcomes. These biases in the data can manifest in various forms, such as overrepresentation or underrepresentation of specific groups, leading to skewed perceptions and decisions made by AI systems.
\\\noindent{}\textbf{Biases in AI Artistic Creation.} Biases in AI artistic creation exemplify how dataset distribution issues can profoundly affect the application of artificial intelligence, often reflecting and even amplifying underlying societal biases. Artistic creation, as an arena where AI has been increasingly applied, offers a stark view into the ramifications of these biases.

When AI systems are employed to process, interpret, or generate artworks from diverse cultural backgrounds or artistic styles, they often manifest clear biases towards certain genres, styles, or races. This phenomenon primarily stems from the composition of the training datasets. If a dataset is heavily skewed towards artworks from a particular cultural or stylistic background, the AI is more likely to develop a bias towards that specific style or culture. This bias can significantly influence the AI’s artistic outputs, subtly shaping the characteristics of the generated artwork in ways that reflect the predominant features of the training data.

The case of GoArt applying an expressionist filter to Clementine Hunter's painting "Black Matriarch" is a telling example\cite{DBLP:journals/corr/abs-2010-13266}. In this scenario, the AI altered the black skin tones to red, a choice that seems inexplicable without considering the training data's influence. In contrast, when processing a sculpture like Desiderio's "Giovinetto," which features a figure with white skin, the AI preserved the artwork’s original color palette. These differences in treatment can be indicative of the AI's learned preferences, potentially influenced by the nature of the training data, which might have contained more frequent representations of certain skin tones within specific artistic contexts.

Such biases in AI artistic creation are not merely academic concerns; they have real-world implications. They can inadvertently perpetuate stereotypes, reinforce cultural hegemonies, and marginalize underrepresented groups. This issue underscores the importance of curating diverse and inclusive datasets, especially in fields like art, where representation and expression are crucial. Furthermore, it highlights the necessity for continuous scrutiny and evaluation of AI models and their outputs to identify and mitigate biases.

In this context, it’s crucial to view AI not just as a technical tool but as an entity shaped by human decisions and societal structures. The choices made in dataset compilation and algorithm design profoundly impact the AI’s behavior, echoing the creators' conscious and unconscious biases. Addressing these issues requires a concerted effort to integrate ethical considerations, cultural sensitivity, and diversity into every stage of AI development, from dataset creation to model training and application.
\\\noindent{}\textbf{Lack of Ethical and Legal Constraints.}
Beyond the challenges posed by statistical distribution, synthetic datasets may also suffer from a lack of necessary ethical and legal constraints during their creation process. This oversight can lead to significant issues, particularly in the context of how data generation algorithms process and interpret the input data. Often, these algorithms may inadvertently learn and replicate biases that are inherent in real-world data sources. This issue is especially pronounced in scenarios involving gender or racial biases.

Moreover, the problem is exacerbated when synthetic datasets rely on publicly available internet data. The internet, as a vast repository of human-generated content, inherently contains a myriad of biases and prejudices that exist within society. This data is often unfiltered and includes implicit societal biases, stereotypes, and even offensive or harmful representations. When such data is used without critical filtering or ethical consideration, the resulting synthetic datasets can inadvertently become a medium through which these societal biases are perpetuated and amplified. The roots of such biases are multifaceted, yet they all converge on a common issue: societal prejudices can infiltrate the process of AI's artistic creation. 

Research\cite{DBLP:journals/corr/abs-2006-16923} indicates that this problem is not limited to synthetic datasets; it also plagues datasets collected from the internet. A prime example is the "Tiny Images" dataset from the Massachusetts Institute of Technology. Compiled using extensive image and label data aggregated from search engines, this dataset was found to contain tendencies of racial and gender discrimination, and even instances of pedophilia, leading to its eventual permanent removal. The emergence of these issues is partly due to the influence of societal biases and partly reflects negligence in the construction of datasets.

In essence, the generation of synthetic datasets requires not only a sophisticated understanding of statistical methodologies but also a deep consideration of the ethical and legal implications. Ensuring fairness and representativeness in these datasets necessitates a comprehensive approach that includes ethical oversight, legal compliance, and an active effort to identify and mitigate any potential biases. This approach should be an integral part of the dataset creation process, ensuring that AI systems trained on these datasets do not perpetuate existing societal biases but rather contribute to fair and equitable outcomes.
% https://www.nature.com/articles/d42473-023-00274-7
% Synthetic data to enhance patient privacy
% https://www.nature.com/articles/s42256-023-00629-1
% Synthetic data accelerates the development of generalizable learning-based algorithms for X-ray image analysis
% https://www.nature.com/articles/d41586-023-01445-8
% Synthetic data could be better than real data

% 4.合成数据集在AI中的风险
% 尽管合成数据在AI应用中发挥了一定作用，但目前的合成数据方法存在一些问题，同时也带来了不合理使用的潜在风险。

% 4.1 合成数据的不足

% 4.1.1 数据分布偏差

% 当前的合成数据方法在处理小样本数据时可能引发数据分布偏差的问题。由于生成的合成数据可能无法充分反映原始数据的分布，导致训练数据的分布与真实数据存在较大的偏差。这种情况下，使用合成数据进行AI训练可能产生模型的偏见，因为模型在学习过程中将受到不真实数据的影响。

% 4.1.2 数据不完整

% 另一个合成数据的不足是生成的数据可能不完整，导致AI模型在预测时出现不准确的结果，从而限制了模型的性能。这可能由于合成数据生成过程中未能捕捉到真实世界数据的复杂性和多样性，导致模型在面对新情境时表现不佳。

% 4.1.3 数据不准确

% 当前大型模型的可解释性存在一定问题，这使得生成的合成数据可能不准确。这样的不准确性可能导致模型发生错误的拟合，使其在实际应用中无法正确处理输入数据。缺乏对生成数据过程的透明性和可解释性可能增加模型的不确定性，使其难以在真实场景中稳健地运行。

% 4.2合成数据在使用中的风险:
% 在实际应用中，使用合成数据有一定的限制和风险，如下:

% 模型通用性能低:大规模使用合成数据将会使得AI模型泛化性受限，如PersonX dataset在是在游戏数据引擎中生成人物身份，和真实世界有较大数据差异；自然语言处理中使用大语言模型微调将导致下游任务受限于选定大语言模型的性能和偏置；医疗领域中，利用合成数据保护了病人的隐私，但利用大量非真实病例训练得到的模型，会使得医生和病人对诊断结果没有信心。

% 潜在的伦理和社会问题:合成数据的使用可能引发潜在的伦理和社会问题。例如，通过合成数据创建虚构的人物或场景，可能会涉及到人工智能在创造虚构内容方面的责任问题。这可能导致误导、误解或者创造虚假的信息，对社会产生不良影响。

% 安全性和对抗性攻击:合成数据的引入也带来了安全性和对抗性攻击的风险。恶意使用合成数据可能导致AI模型在面对对抗性攻击时表现不稳定，因为模型在训练过程中未必能够有效地学习真实世界的复杂性和多样性。这可能使得模型更容易受到欺骗或误导，对系统的可信度和安全性构成威胁。

% 法律合规性:在某些领域，合成数据的使用可能涉及法律合规性的问题。例如，在金融领域使用合成数据进行风险评估，可能面临监管方面的挑战，因为监管机构通常要求透明和可解释的模型，而合成数据的生成过程可能难以满足这些要求
\section{Risks and Challenges in Utilizing Synthetic Datasets for AI.}
While synthetic data plays a crucial role in AI applications, the current methods of generating synthetic datasets bring forth notable challenges and potential risks, necessitating careful consideration of their applications.
\subsection{Shortcomings of Synthetic Data.}
\noindent{}\textbf{Data Distribution Bias.} A discernible incongruity exists between synthetic datasets and their authentic counterparts, encompassing notable disparities in feature distribution, class distribution, and other pertinent statistical attributes. This bias imparts a proclivity for models to engender misleading prognostications within practical applications, thereby compromising their fidelity to faithfully encapsulate real-world phenomena.
\\\noindent{}\textbf{Incomplete Data.}The presence of lacunae or partial information within synthetic datasets, ostensibly stemming from imperfections, errors, or an inadequacy in encapsulating the manifold changes inherent in authentic datasets during the synthetic generation process. This dearth of comprehensive information may impede the model's capacity to accurately prognosticate or manage scenarios characterized by data incompleteness, thereby influencing the model's resilience and pragmatic utility.
\\\noindent{}\textbf{Inaccurate Data.}
The manifestation of errors, noise, or inaccuracies within synthetic datasets, diverging significantly from the veracity of real-world datasets. This discrepancy may arise from algorithmic imperfections, noise injection, or other contributory factors that give rise to inaccuracies. Consequently, the model may internalize erroneous patterns, thereby inducing biased predictions and undermining the overall performance and reliability of the model when confronted with genuine data.
\\\noindent{}\textbf{Insufficient Noise Level.}
 Synthetic datasets may evince an undue sterility, lacking the multifarious noise and intricacies inherent in real-world data. In authentic scenarios, data invariably incorporates diverse interferences, errors, and uncertainties. The paucity of such features within synthetic datasets may hamper the model's efficacy within realistic environments.
\\\noindent{}\textbf{Over-Smoothing.}
In the process of synthetic data generation, certain models may overly attenuate or oversimplify the data, resulting in an attenuated representation devoid of the nuanced details and diversity inherent in authentic datasets. This propensity may precipitate challenges for the model in assimilating complex variations within genuine data.
\\\noindent{}\textbf{Neglecting Temporal and Dynamic Aspects.}
 Certain methodologies for synthetic data generation may inadequately capture temporal and dynamic nuances, facets that are inherently pivotal within authentic datasets. The consequential failure to faithfully simulate these temporal intricacies may culminate in the ineffectuality of models in real-world applications.
\\\noindent{}\textbf{Inconsistency.}
 The paucity of inconsistency within synthetic datasets, relative to the rich tapestry inherent in authentic datasets. The latter frequently embodies variations stemming from diverse sources, temporal epochs, and environmental conditions, aspects that synthetic datasets often fail to encapsulate. This shortfall may engender challenges for models in adapting to the multifaceted vicissitudes originating from disparate sources, temporal epochs, and environmental conditions, thereby precipitating a decrement in the generalization performance vis-à-vis diverse datasets.

\subsection{Risks in Synthetic Data Application.}
\noindent{}\textbf{General Model Performance}
Widespread use of synthetic data may constrain the generalization performance of AI models. For instance, datasets like PersonX, generated within a gaming data engine, may deviate significantly from real-world data. In natural language processing, relying on fine-tuning with large language models may restrict downstream tasks to the performance and biases of the selected model. In healthcare, an abundance of non-real cases during model training may undermine confidence in diagnostic results among healthcare professionals and patients.
\\\noindent{}\textbf{Ethical and Social Implications}
The use of synthetic data may give rise to ethical and social concerns. Creating fictional characters or scenarios through synthetic data raises questions about AI responsibility in generating fictional content, potentially leading to misinformation, misunderstandings, or the dissemination of false information with detrimental societal impacts.
\\\noindent{}\textbf{Security and Adversarial Risks}
The introduction of synthetic data brings forth security and adversarial attack risks. Malicious use of synthetic data may render AI models unstable during adversarial attacks, as models may not adequately learn the complexity and diversity of the real world during training. This susceptibility increases the likelihood of deception or manipulation, posing threats to the credibility and security of the system.
\\\noindent{}\textbf{Legal Compliance Challenges}
In certain domains, the use of synthetic data may present challenges regarding legal compliance. For instance, employing synthetic data for risk assessment in the financial sector may encounter regulatory hurdles. Regulatory authorities often require transparent and interpretable models, and the synthetic data generation process may face difficulties in meeting these standards.

% 综合而言，合成数据集在AI中的使用带来了一系列潜在风险，需要谨慎评估和管理。合成数据生成方法需要更加精细和逼真，以减少与真实数据之间的差异。在实际使用中，对于合成数据和真实数据的混合使用，以及对模型的进一步验证和测试，都是减缓这些风险的有效手段。只有充分意识到这些潜在问题，我们才能更好地利用合成数据来训练鲁棒性更强、更可靠的人工智能模型。
% 综合而言，合成数据的不足可能会影响AI模型的性能和可靠性，需要在合成数据的生成和使用中加强监管和规范，以确保生成的数据质量和适用性。
\section{Conclusions}
In summary, the use of synthetic datasets in AI introduces a range of potential risks that require careful assessment and management.
\subsection{New Approaches to Synthetic Data.}
To address the current issues associated with synthetic data generation, there is a need for continuous development and adoption of new methods. 
\\\noindent{}\textbf{Adopting more advanced generative models. }One potential approach involves the use of advanced generative models such as Generative Adversarial Networks (GANs) or Variational Autoencoders (VAEs). These models possess stronger learning capabilities, allowing for more accurate modeling of the complex distribution of real-world data. By employing these advanced models, it becomes possible to better avoid distribution shift issues, enhance the diversity of generated data, and more effectively simulate the noise and uncertainty present in the real world.
\\\noindent{}\textbf{Integrating domain-specific expertise to enhance the realism of synthetic data. }Integrating domain-specific knowledge, such as computer graphics, physics, and cognitive science, can contribute to improving the realism of synthetic data. A deeper understanding of the physical laws behind scenes and cognitive processes can lead to more precise generation of various scenarios, making synthetic data closer to real-world situations.
\subsection{Regulating the Use of Synthetic Data.}
To regulate the use of synthetic data, establishing a set of clear guidelines is crucial. \\\noindent{}\textbf{Establishing industry standards. }Industry standards should be developed to outline best practices for synthetic data generation, covering the selection of data generation models, parameter settings, and the correlation between synthetic and real data.
\\\noindent{}\textbf{Transparency and documentation. }Transparency and documentation are equally essential. Researchers and practitioners should clearly document the methods and parameter settings used for generating synthetic data, providing detailed documentation about the synthetic data set. This aids other researchers in understanding the source and characteristics of the data, facilitating a better assessment of the model's performance.
\\\noindent{}\textbf{Emphasizing model validation and evaluation. }Emphasizing model validation and evaluation is a crucial step in regulating the use of synthetic data. In addition to training on synthetic data, models should be validated on real data to ensure their generalization performance and robustness. Regularly updating synthetic data sets to adapt to new scenarios and changes in data distribution is also a vital means of maintaining model performance.
%数据产生方法，
%数据使用情况
%两方面介绍关于合成数据集在AI中使用问题
%建议:
%5.1新的方法合成数据
%5.2规范合成数据的使用
\bibliographystyle{ieeetr}
\bibliography{reference}

\end{document}